\documentclass{article}

\usepackage{arxiv}

\usepackage[utf8]{inputenc} 
\usepackage[T1]{fontenc}    
\usepackage{hyperref}       
\usepackage{url}            
\usepackage{booktabs}       
\usepackage{amsfonts}       
\usepackage{nicefrac}       
\usepackage{microtype}      
\usepackage{lipsum}		
\usepackage{graphicx}
\usepackage{doi}
\usepackage{tikz}
\usepackage{comment}
\usepackage{amsmath,amssymb} 
\usepackage{color}
\usepackage[bottom]{footmisc}
\usepackage{footnote}
\usepackage{listings}
\usepackage{xcolor}
\usepackage{subcaption}

\title{Crowd Source Scene Change Detection and Local Map Update}

\author{
    Itzik Wilf\thanks{Authors contributed equally to this work.}\\
    Toga Networks\\
    \texttt{itzik.wilf@toganetworks.com} \\
    \And
    \href{https://orcid.org/0000-0002-0939-3379}{\includegraphics[scale=0.06]{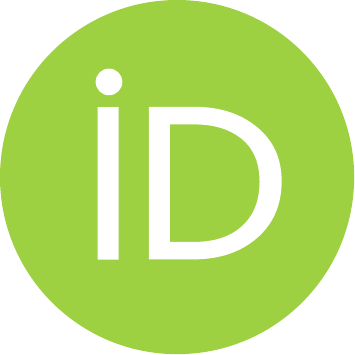}} Nati Daniel\footnotemark[1] \\
    Toga Networks\\
    \texttt{Nati.Daniel@toganetworks.com} \\
    \And
    Lin Manqing\footnotemark[1] \\
    Huawei\\
    \texttt{linmanqing@huawei.com} \\
    \And
    Firas Shama \\
    Technion\\
    \texttt{shama.firas@gmail.com} \\
    \And
    Omri Asraf \\
    Toga Networks\\
    \texttt{omri.asraf1@toganetworks.com} \\
    \And
    Feng Wensen\thanks{Co-corresponding authors.} \\
    Huawei\\
    \texttt{fengwensen@huawei.com} \\
    \And
    Ofer Kruzel\footnotemark[2]\\
    Toga Networks\\
    \texttt{ofer.kruzel@toganetworks.com} \\
}

\date{}

\hypersetup{
pdftitle={Crowd Source Scene Change Detection and Local Map Update},
pdfsubject={q-bio.NC, q-bio.QM},
pdfauthor={Itzik.~Wilf, Nati.~Daniel, Lin.~Manqing, Firas.~Shama, Omri.~Asraf, Feng.~Wensen, Ofer.~Kruzel},
pdfkeywords={Scene Change Detection,Map Update,Visual Positioning System (VPS),Image Segmentation,Image Pair Alignment,3D Point Cloud,Deep Learning},
}

\begin{document}
\maketitle

\begin{abstract}
	As scene changes with time map descriptors become outdated, affecting VPS localization accuracy. In this work, we propose an approach to detect structural and texture scene changes to be followed by map update. In our method - map includes 3D points with descriptors generated either via LiDAR or SFM. Common approaches suffer from shortcomings: 1) Direct comparison of the two point-clouds for change detection is slow due to the need to build new point-cloud every time we want to compare; 2) Image based comparison requires to keep the map images adding substantial storage overhead. To circumvent this problems, we propose an approach based on point-clouds descriptors comparison: 1) Based on VPS poses select close query and map images pairs, 2) Registration of query images to map image descriptors, 3) Use segmentation to filter out dynamic or short term temporal changes, 4) Compare the descriptors between corresponding segments.
\end{abstract}


\section{Introduction}
3D point clouds reconstructed from image-based Structure-from-Motion (SfM) or RGB and LiDAR are often frozen in time and thus gradually loses its ability to model the constantly changing environment with high fidelity. The first step towards maintaining an up-to-date city-scale 3D model is to detect changes in the geometric structure as well as the textures of the scene, while excluding other nuisance factors such as appearance changes from illumination, viewpoint differences or short term changes due to people, sky or cars. Detecting temporal changes in a city is an important problem, with many applications such as maintaining updated maps for autonomous driving systems, AR/VR and robotics applications.

One naıve way of computing the changes is to directly compare images between the two traversals using some variant of image differencing~\cite{thresholding,mahmoudzadeh2007digital}. However, such approaches require to hold the map images with the map and are sensitive to illumination differences between the two acquisitions. In addition, they require near pixel-perfect alignment to work well which can be hard to achieve on a moving camera. Several works \cite{alcantarilla2018street,taneja2011image} tackle this problem using a 3D model of the scene. These typically reconstruct a dense 3D model and recover the camera poses using SfM and multi-view stereo (MVS) techniques. The dense models can later be used for dense image alignment \cite{alcantarilla2018street} or to reproject pixels between images to detect inconsistencies \cite{taneja2011image}. However, obtaining accurate camera poses across traversals can be challenging. VPS errors can often be several meters in urban environments; SfM techniques can also fail due to appearance differences which can be due to illumination differences or even large scene changes. In addition,reconstructing the dense model is computationally expensive and dense reconstructions are arguably unnecessary for many applications such as localization of autonomous vehicles \cite{shi2019visual} robotics and AR/VR.

To avoid these problems, we register and detect changes using the 3D point clouds against their corresponding 2D features within the query image.

Experiments show our approach can detect these structural and texture changes even in the presence of large viewpoint and illumination differences.\\
To summarize, our contributions are as follows:
\begin{itemize}
    \item Query to Map Geometric Image Pair Selection (QMGIPS) using query images poses from VPS service.
    \item Aligning QMGIPS pairs to support local comparison of appearance by 5DOF Homography RANSAC estimation algorithm.
    \item Robust change detection under appearance changes.
    \item Semantic segmentation filtering of short term and exempt objects.
    \item Visual segmentation based visual features change detection.
    \item Indirect map change detection by extending direct change map image to All map images via Image Retrieval algorithm.
    \item Efficient approach for local map update, which shows better localization accuracy, and more faster run time.
\end{itemize}

\begin{figure}
\centering
\includegraphics[width=\textwidth]{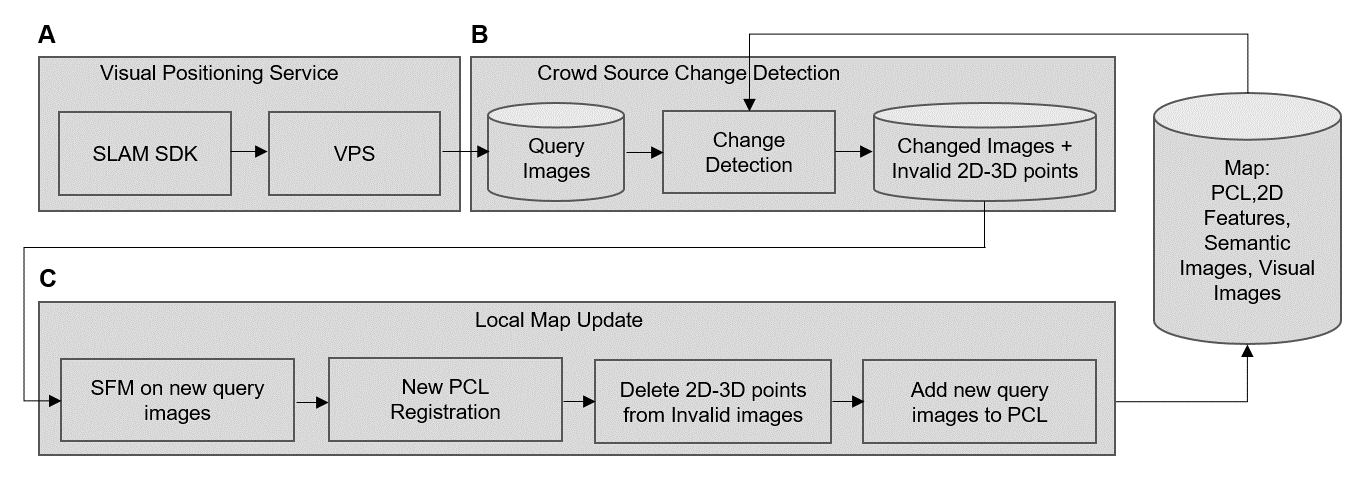}
\caption{\textbf{Crowd Source Change Detection and Local Map Update - High Level Description}. A. VPS query images stored on server side for further processing. B. Crowd Source Change Detection service periodical run and detect map changes and store changes on server including change location. C. Local Map Update service shall update the map. PCL, 3D Point Cloud. SFM, Structure From Motion.}
\label{fig:f1}
\end{figure}

\section{Related Work}
\subsubsection{Change Detection.} Change detection methods can be broadly classified as 2D or 3D. 2D methods generally compare input images though some variant of image differencing \cite{thresholding,mahmoudzadeh2007digital}, and tend to be sensitive to illumination differences or misalignments between the images. To partially overcome these issues, the images are often preprocessed to remove illumination variations \cite{radke2005image}, and registered \cite{brown1992survey} to each other. Despite this, 2D methods remain sensitive to viewpoint changes as they typically require pixel perfect registration to work well. Also, most 2D methods detect appearance changes which may not correspond to actual scene changes. 3D methods make use of a known 3D scene structure or reconstruct it from the input images to better detect structural changes. Taneja et al. \cite{taneja2011image} assumes that a 3D model of the scene in the previous time step is available and detects changes by checking the consistency during projection between images in the later time step. Ulusoy and Mundy \cite{ulusoy2014image} extends this to infer changes in the 3D model itself. Sakurada et al. \cite{sakurada2013detecting} foregoes the dense model and instead makes use of stereo pairs in both time steps to perform the reprojection. Another direction is to generate a spatio-temporal model from images captured at various times by incorporating time into SfM methods. \cite{schindler2010probabilistic,matzen2014scene} infer the temporal ordering of images and the temporal extent of 3D points in the scene by analyzing the SfM output. Lee and Fowlkes \cite{lee2017space} optimizes a probabilistic spatial-temporal model using expectation maximization to simultaneously register 3D maps and infer the temporal extents of scene surfaces. Most of the above methods require accurate relative camera poses between the two times, which can be difficult to obtain especially when the scene has changed. 

\subsubsection{Change Detection using Deep Learning.} More recently, several works \cite{alcantarilla2018street,guo2018learning,varghese2018changenet} apply deep learning to detect scene changes. These works learn to compare two input images to detect changes in a strongly supervised manner, requiring pixel level \cite{alcantarilla2018street,guo2018learning,zhan2017change,varghese2018changenet} or patch level \cite{stent2015detecting} annotations which can be highly tedious to obtain. To avoid the need for annotation, Sakurada and Okatani \cite{sakurada2015change} compare normalized features extracted from an upper layer of a convolutional neural network pretrained on a image recognition task, and make use of super-pixel segmentations to obtain high resolution outputs. Despite the good performance shown by these works, these image-based methods remain sensitive to viewpoint differences since they do not consider the 3D structure of the scene. Alcantarilla et al. \cite{alcantarilla2018street} partly alleviates this issue by performing a dense warp between images, but requires a computationally expensive dense reconstruction and accurate relative camera poses across the two traversals. 

\subsubsection{Point Cloud Registration.} The above change detection works often require to keep the map images with the map causing additional storage overhead as well as accurate image registration, which can be difficult to achieve under scene changes or illumination variations. We circumvent these challenges by comparing only the point cloud features that exist in the map to the features in the query image. Point cloud registration methods can be broadly classified into 1) feature-based methods \cite{rusu2009fast,salti2014shot} which establish correspondences by matching descriptors before computing the transformation, and 2) simultaneous pose and correspondence methods \cite{besl1992method,chen1992object} that typically use iterative schemes to estimate both pose and correspondences. Learned variants of both feature-based \cite{zeng20173dmatch,choy2019fully} and simultaneous pose and correspondence \cite{aoki2019pointnetlk,yew2020rpm} methods are also available. DeepMapping \cite{ding2019deepmapping} optimizes for the registration objective by training a neural network. Depending on the application, point clouds may undergo local deformations which requires estimating a nonrigid transformation. In this work, we take a new approach and use only existing map point cloud features and compare them to features in query images to detect bot structural and texture changes.

\section{Approach}
Herein, we propose an effective pipeline for map change detection and update, which is fully automatic and based on crowd source queries – merely whenever the map is used for localization, as shown in Fig.~\ref{fig:f1}. As the process is accumulating evidence of change before updating the map, it is best implemented and described as a batch process running on a sequence of query images and the computed poses. Blocks Fig.~\ref{fig:f1}A and Fig.~\ref{fig:f1}B, detect changes by comparing query images to the map, and then remove obsolete 2D-3D points from the map. Step Fig.~\ref{fig:f1}C performs map augmentation by creating a new point cloud at the change area, registering it with the map and adding the new images 2D-3D points to the map.

\begin{figure}
\centering
\includegraphics[width=\textwidth]{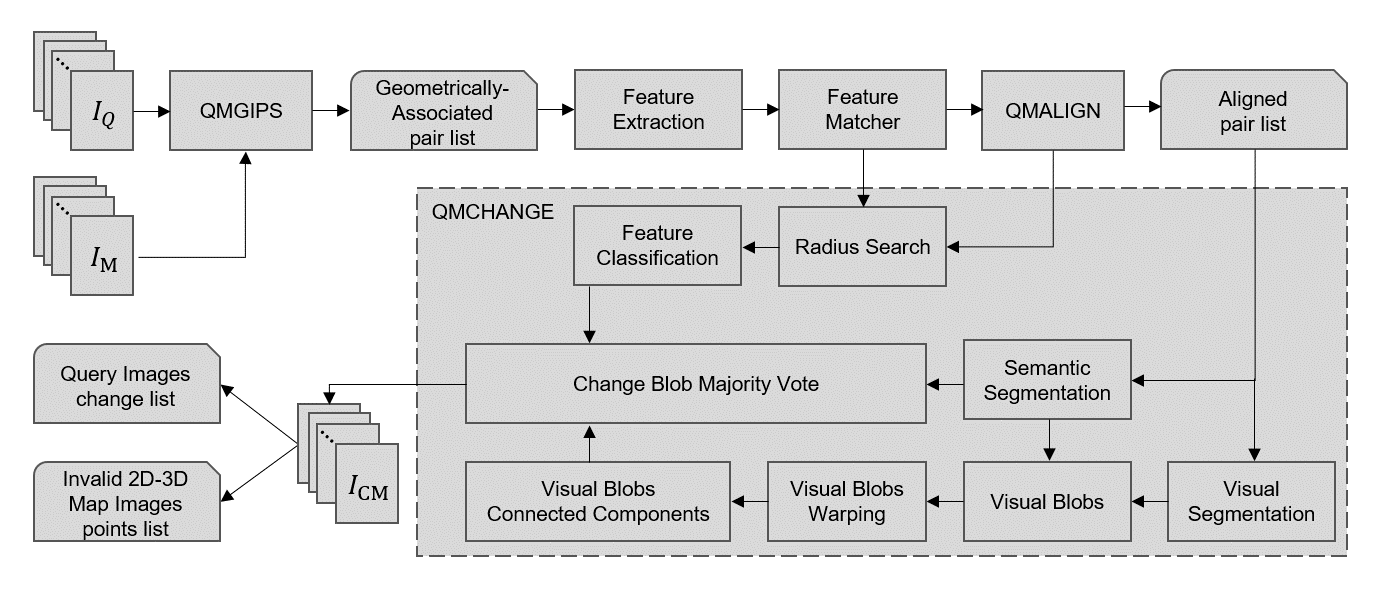}
\caption{\textbf{Image Pair Change Detection Pipeline}. Our Crowd Source Change Detection mechanism is based on a novel Appearance and Location approach. Where $I_{Q}$ indicates query images, $I_{M}$ indicates map images, and $I_{CM}$ indicates changed mask images.}
\label{fig:f2}
\end{figure}

\subsection{Map Change Detection.}
Our pipeline comprises the following steps as described in Fig~\ref{fig:f2}:
\begin{itemize}
    \item Query-Map Geometric Image Pair Selection (QMGIPS): a fast process of comparing the pose of a query image against all map images poses and selecting query-map image pairs that can be reliably compared for change detection. Given such pair, we can move to the next step.
    \item Query-Map Image Alignment (QMALIGN): Change detection is a local process in which corresponding portions of the images are compared. Therefore, image alignment is required. QMGIPS pairs can be effectively aligned by homography. We have developed a 5DOF homography alignment process which is usually more robust than the customary 8DOF process. We also developed a homography refinement step which increases the number of inliers and the perceived quality of alignment. QMALIGN also serves as a verification to QMGIPS, and once successful (as judged by the number of inliers) we proceed to the next step.
    \item Query-Map Change Detection (QMCHANGE):  given an aligned pair, change detection is done by comparing local image descriptors.  Each  map image keypoint descriptor is compared against nearby query keypoint image descriptors (transformed by homography) in semantically relevant image locations. Matching failure is an indication of change. However to make such matching robust and generate change masks from distinct keypoints, we use visual segmentation as support \cite{DIC}. Warping query image visual segments onto the map image, we count changed and unchanged features. We classify such a segment as changed/unchanged based on change ratio/majority voting.
\end{itemize}

\subsubsection{Geometric Image Pair Selection (QMGIPS).}
Most image change detection algorithms \cite{guo2018learning,sakurada2020weakly,yew2021city} compare images captured from the same position (e.g. surveillance cameras). This would make both images related by homography (2D-2D perspective transform) and facilitate comparison. However, this is not the case in VPS with different camera locations of the query image and the relevant map images. Different query-map camera locations result in parallax which make image matching and change detection difficult. Our experience with techniques like RANSAC-FLOW \cite{ransacflow}, which uses two-stage for calculating fine flow for image alignment. The first stage use deep features to calculate piecewise homographies for coarse alignment and after that use a CNN in order to calculate local fine flow, did not produce good results. Moreover, some elements seen in one image are invisible in the second image, simply because of viewpoint change and might be falsely perceived as scene changes. Therefore, we choose very specific pairs of query-map images to compare. Classical approaches for image pair selection, (1) Sequential, (2) Visually Similar where, visual similarity is error prone in repeated texture. Our solution, the Query-Map Image Pair Selection (QMGIPS) algorithm, is to use SLAM poses and camera Field of View (FOV) to verify that similar images depict same scene. In addition to reducing parallax and facilitates alignment. QMGIPS compares the location and orientation of each query image against ALL map images, selecting pairs with camera locations no farther than $D=1$ meters apart and line-of-sight (LOS) directions no farther than $\Delta\theta=0.2$ radians apart. The large number of map images times the large number of received VPS queries, is supposed to generate a sufficient number of image pairs to compare.

\begin{figure}
\centering
\includegraphics[scale=0.5]{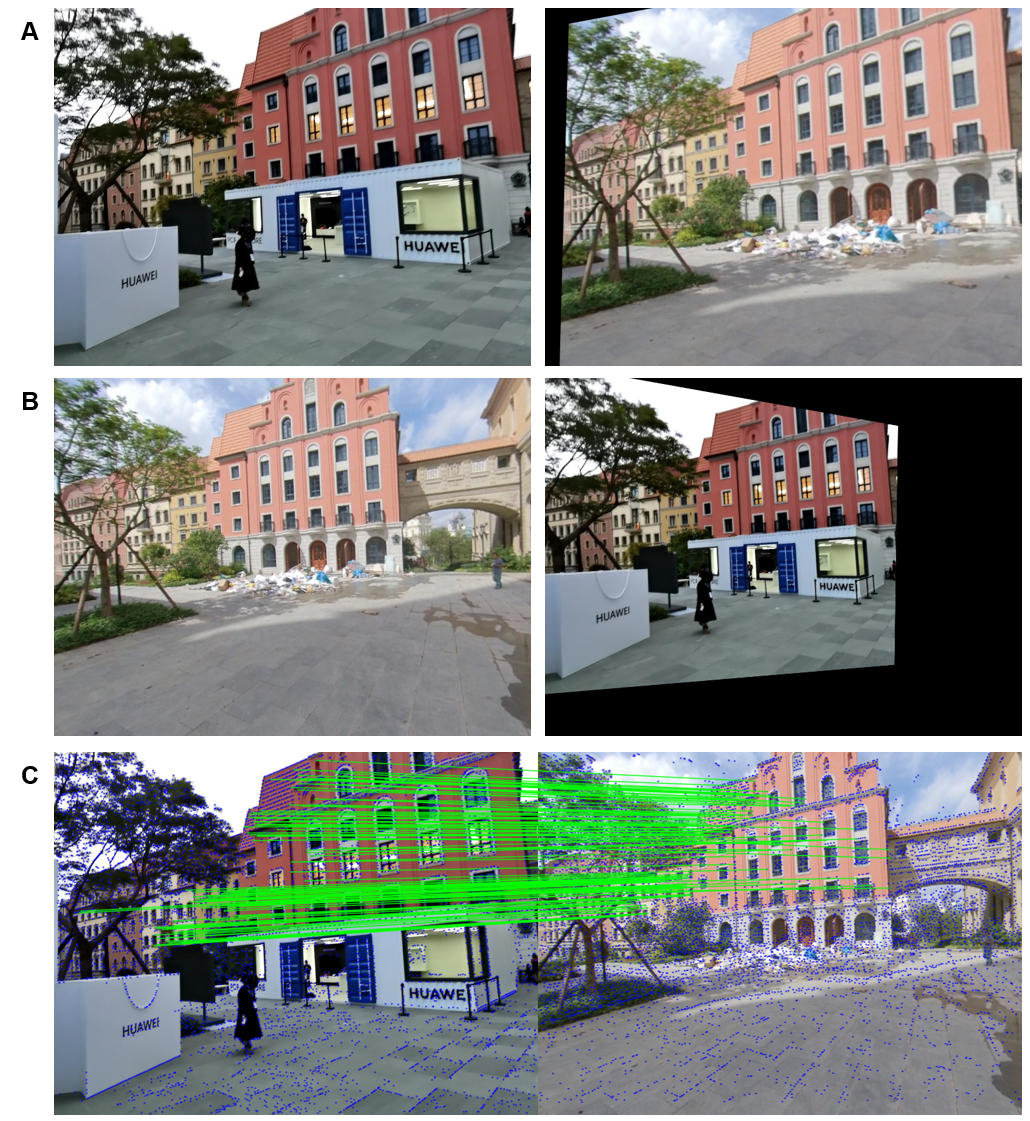}
\caption{\textbf{5DOF RANSAC Homography Pair Alignment}. A. A query image left, map aligned image right. B. A map image left, query aligned image right. C. Bi-Directional Good Matches of 2D D2Net features (blue dots), which used for homography estimation. Black part of the images indicates the non-overlay scenes between the query and map images. Images are part of D457 dataset.}
\label{fig:f3}
\end{figure}

\subsubsection{Image Pair Alignment (QMALIGN).}
QMGIPS image pairs can be aligned quite well by a single homography. Planar homography has 8 degrees of freedom (8DOF), can be computed from $4$ matching image point pairs.

Given a large number of image matches, we can use the RANSAC paradigm to prune outliers and compute homography on the remaining matches. However, in reality 8DOF homography frequently results in titled planes and other distortions. 
Hence, under the assumptions that images are captured essentially from the same height (relative to the scene dimensions) with the camera in upright position, not rotated or titled. It is clearly observed by buildings’ vertical parallel lines remaining vertical and parallel under perspective. Perspective foreshortening is visible along the horizontal (X) axis only. Therefore, we can use a degenerate case of homography, one with 5 degrees of freedom (5DOF), which preserves vertical lines. Herein, we have developed a 5DOF RANSAC algorithm for Image pair alignment (see the supplementary material, Algorithms section).

The metrics with which we evaluated the 5DOF Alignment are the number of RANSAC inliers, and $\sigma_{x,y}$. When, number of RANSAC inliers indicates the confidence of how the alignment is accurate, and $\sigma_{x,y}$, which is the standard deviation of the homography inliers, indicates how the inliers scatter along the image. As large $\sigma_{x,y}$ as better spatial distribution of the inliers within the image. In particular, we used a threshold of minimum $80$ inliers for a successful pair alignment, while the $\sigma_{x,y}$ indications are used for debugging.

The image pair in Fig.~\ref{fig:f3} is a difficult example, with uncorrected orientation (right), and significant changes. Yet, D2Net features \cite{d2net} are nicely paired by bidirectional matching. The 5DOF RANSAC algorithm aligns the corresponding parts of both images, preparing them for Image Pair Change Detection (CD).

\subsubsection{Map Change Detection Pre Process.}\label{sec:PreProcess}
\begin{figure}
\centering
\includegraphics[scale=0.6]{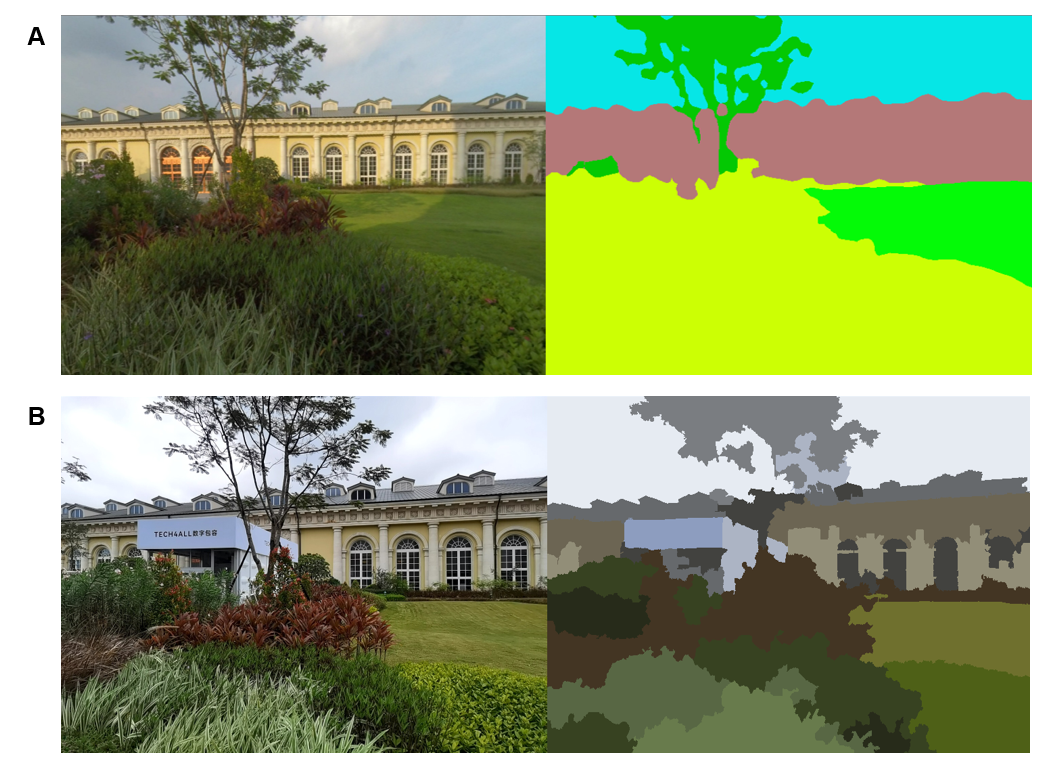}
\caption{\textbf{Deep image segmentation: Semantic Segmentation and Visual Segmentation}. A. Map Image and its Semantic Segmentation. B. Query Image and its Visual Segmentation.}
\label{fig:f4}
\end{figure}

\begin{figure}
\centering
\includegraphics[scale=0.45]{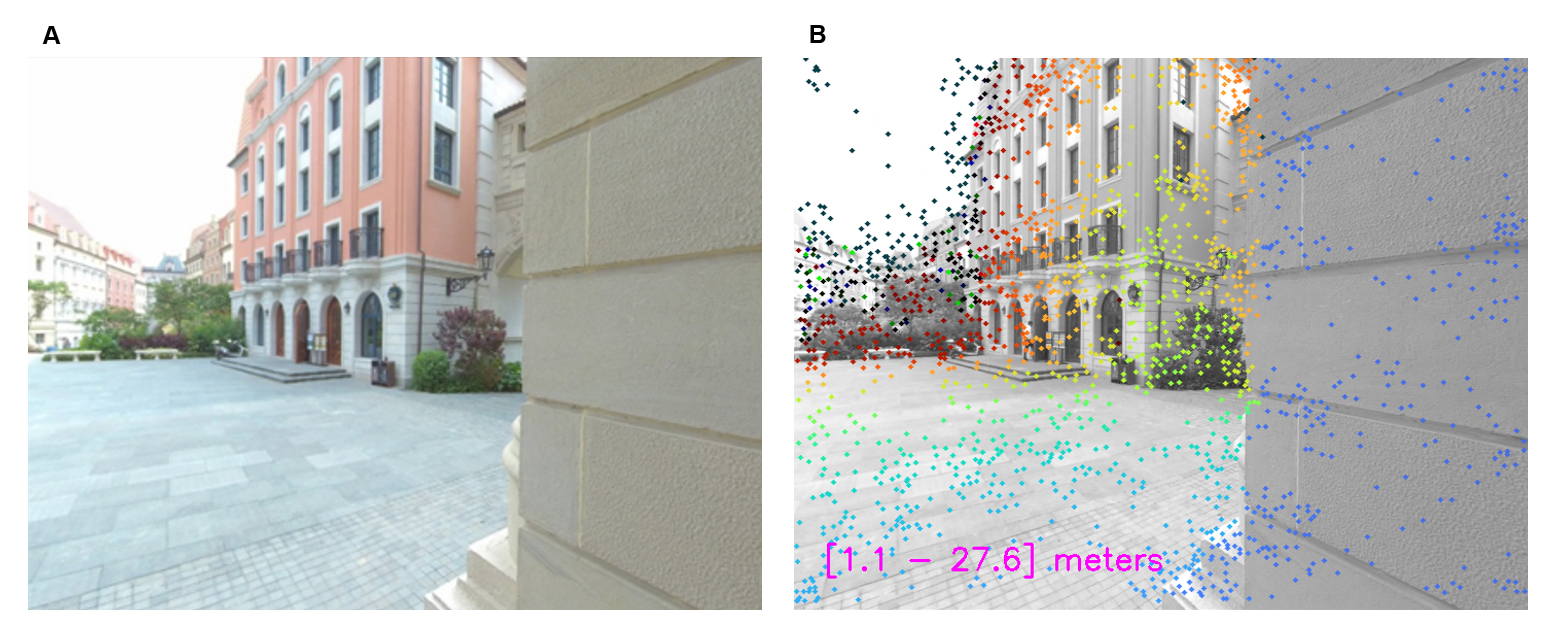}
\caption{\textbf{Depth Filtering}. A. A typical Map image of D457 dataset. The image was taken at high resolution of 1440X1080 pixels. B. A Map images with extracted D2Net features \cite{d2net} with their relative distance to the camera pose. Blue features points are closer to the camera up to 3 meters, Black features points have no 3D information. The text marked in purple color indicates the range of the color Coding Distance from Camera.}
\label{fig:f5}
\end{figure}

The pre-processing stage includes applying semantic and visual segmentation equally to the query, and to each relevant map images for the next stages, and filtering close objects that conceal the view. 
\begin{itemize}
    \item Semantic segmentation is useful for focusing our change detection on image classes of interest – e.g. ignoring sky, plants and trees. However, as you can see above, the structure class (in brown) covers a very large area and does not yield fine-grain regions for change detection (as shown in Fig.~\ref{fig:f4}A). For this purpose, we use DeepLab model \cite{DeepLab_semantic} with ResNetSt269 backbone \cite{ResNeSt_backbone_for_semantic} as provided by PyTorch Encoding package \cite{PyTorchEncoding}. This model provides SOTA solution for the ADE20K dataset \cite{zhou2016semantic_ade20k_1,zhou2017scene_ade20k_2}, which benchmark indoor scenarios, and relevant to our scenario. ADE20K dataset contains images of indoor and outdoor scenes, includes scene segmentation, object detection, and 150 different labels.
    \item Visual segmentation breaks the images into fine regions with visual coherence. Such regions will be used as the basic elements of change detection (as shown in Fig.~\ref{fig:f4}B). For this purpose, we use Deep Image Clustering for Unsupervised Image Segmentation (DIC) model \cite{DIC}. DIC is an unsupervised segmentation framework based on a novel deep image clustering (DIC) model. The DIC consists of a feature transformation subnetwork (FTS) and a trainable deep clustering subnetwork (DCS) for unsupervised image clustering. FTS is built on a simple and capable network architecture. DCS can assign pixels with different cluster numbers by updating cluster associations and cluster centers iteratively. Moreover, a superpixel guided iterative refinement loss is designed to optimize the DIC parameters in an overfitting manner.
    \item Our QMGIPS strategy assumes that the viewpoint distance between the poses of the query-map pair is small compared with the distance from the cameras to the respective scene features. Scene elements close to the camera may violate that assumption and appear as false change after alignment. To avoid that, we compute the distance of features from the camera and simply ignore features closer than a pre-specified distance (3.0m) (as shown in Fig.~\ref{fig:f5}).
\end{itemize}

\subsubsection{Query-Map Change Detection (QMCHANGE).} \label{sec:QMCHANGE} 
In this section, we focus on the change detection problem - given a pair of images with an overlapping area, make a mask for each image that covers the changes (see Fig. \ref{fig:f3}.C). The proposed QMCHANGE algorithm could be describe by three main blocks: pre-processing, features map decision and decision over local segment areas (blobs).

After the pre-processing (see Section~\ref{sec:PreProcess}), the next stage is building a feature map by classifier for each feature if it has a good matching in the other image by two different conditions: visual (descriptor distance) and geometry (radius search). The visual threshold we defined to be a constant multiply median of descriptor distance of all the matching we got by opencv K-Nearest Neighbors (KNN) feature matcher. The geometry threshold we defined to be maximum 50 pixel distance between the features after alignment transformation.

 The next step is taking decision in patch level by change blob majority vote. First based the semantic segmentation we filtered all the labels that we not consider as a change like: sky, ground, humans and plants. Second, based the alignment result we filter all the image areas that not in the line of sight in the corresponding image. Next, using the visual segmentation labels we divided the image to blobs. In each blob we calculate the ratio between the amount of "good features" to the "bad features". A blobs where the ratio of "bad features" be higher then a threshold will considered as a change. By connecting all the local changes we will get the changes mask. 
 
 Finally, for more clean mask we delete all small blobs and fill small holes. Note all the change detection process should be done twice, one time on map image and second on the query image.

\subsection{Local Map Update}
With time, scene changes affect the relevance of the map, reducing the success rate and accuracy of subsequent localization. So, a map update is required. Robust, but inefficient approaches to map update include either capturing a new map (relevant portion of the scene) periodically, or when being notified on change by authorities, human notification, business owners, or users. \\
Our Map update approach is fully automatic and based on crowd source queries – merely whenever the map is used for localization. As the process is accumulating evidence of change, the stored elements are incrementally added to the current map in an SFM procedure (AKA Map Augmentation). Then the changed 2D-3D features in all affected images are deleted from the map (global descriptors, cloud points with associated key points and local descriptors).

\subsubsection{Change aggregation and change propagation (QMPOST).}
A QMGIPS pair processed by QMALIGN and QMCHANGE generates a change mask, which can be quite noisy. Our QMPOST process aggregates change masks per map image. Averaging a sufficient number of such masks, we obtain a robust map image change detection mask. However, only a fraction of map images make it to final robust mask step, while a much larger number of map images depict these changes. We propagate the change from the “Master” map images to other images, using visual similarity and spatial proximity \cite{netvlad} (a pretrained CNN encoder, followed by a learnt VLAD component, PCA and whitening). Propagation marks changed 2D-3D features in all affected images, as such. When localizing future query images against the map, we ignore 2D-3D features marked as changed. For more details, see the supplementary material, Experiments section.
\subsubsection{Map Augmentation.}
Although our pipeline detects changes and removes changed features from the map, with time we can expect the map to become sparser. Future localization will have lower success and accuracy rate. Eventually we need to augment the map with new data. Our change propagation process marks the changed map images and the extent of change, so we can remap only the relevant locations by running SfM procedure. Given a new map portion, we align it with the old map, using 3D-3D RANSAC (see the supplementary material, Algorithms section) followed Iterative Closest Point (ICP) \cite{arun1987,besl1992,chen1992,zhang1994}.

\section{Experiments}
\subsection{Datasets}
For evaluation purposes we chose two outdoor datasets, commonly used for benchmarking Map Change Detection: Business District (BD) and Research Town (RT) \cite{yew2021city}, and two more outdoor datasets, denoted as D285 and D457, for benchmarking Map Change Detection and Update. The BD and RT datasets include 30 image pairs each, which were randomly selected. Each pair includes 2 similar viewpoint points that were taken at two different times $[t_{0},t_{1}]$, in order to achieve different illumination variations and structural changes in the scenes. While the D285 and D457 datasets were constructed directly for this research with the help of the Huawei’s Shenzen research center people.
All these datasets encompass common challenges in scene change detection, and in particular in scenarios typical of large scale visual localization applications: indoor/outdoor, occlusions, repeating textures, significant viewpoint changes, pair alignemnt is not perfect, and camera trajectory variations between query and map images. 
The main parameters of all four datasets are given in Table~\ref{tab:datasets}. We report results on all these datasets in Section~\ref{sec:Results}.

\begin{table}
    \centering
	\caption{Main parameters of all four datasets used in Map Change Detection and Update processes.}
	\label{tab:datasets}
	\setlength{\tabcolsep}{2.2pt}
	\begin{tabular} { |l|c|c|c|c| }
	        \hline
			\textbf{Parameter/Dataset} & \textbf{BD}~\cite{yew2021city} & \textbf{RT}~\cite{yew2021city} & \textbf{D285} & \textbf{D457}\\
			\hline
			\# of map images & $15$ & $15$ & $17,652$ & $13,092$\\			
			\hline
			\# of query images & $15$ & $15$ & $1,945$ & $1,374$\\
			\hline
			\# of VPS images & $-$ & $-$ & $972$ & $687$\\
			\hline
			Original image size & $1642X1370$ & $1642X1370$ & $1440X1080$ & $1440X1080$\\
			\hline
			Crop image size  & $821X685$ & $821X685$ & $720X540$ & $720X540$\\
			\hline
			Groundtruth change mask & Yes & Yes & No & No\\
			\hline
			Given pairs & Yes & Yes & No & No\\
			\hline
			Pair alignment & Yes & Yes & No & No\\
			\hline
	\end{tabular}
\end{table}

\begin{figure}
\centering
\includegraphics[width=\textwidth]{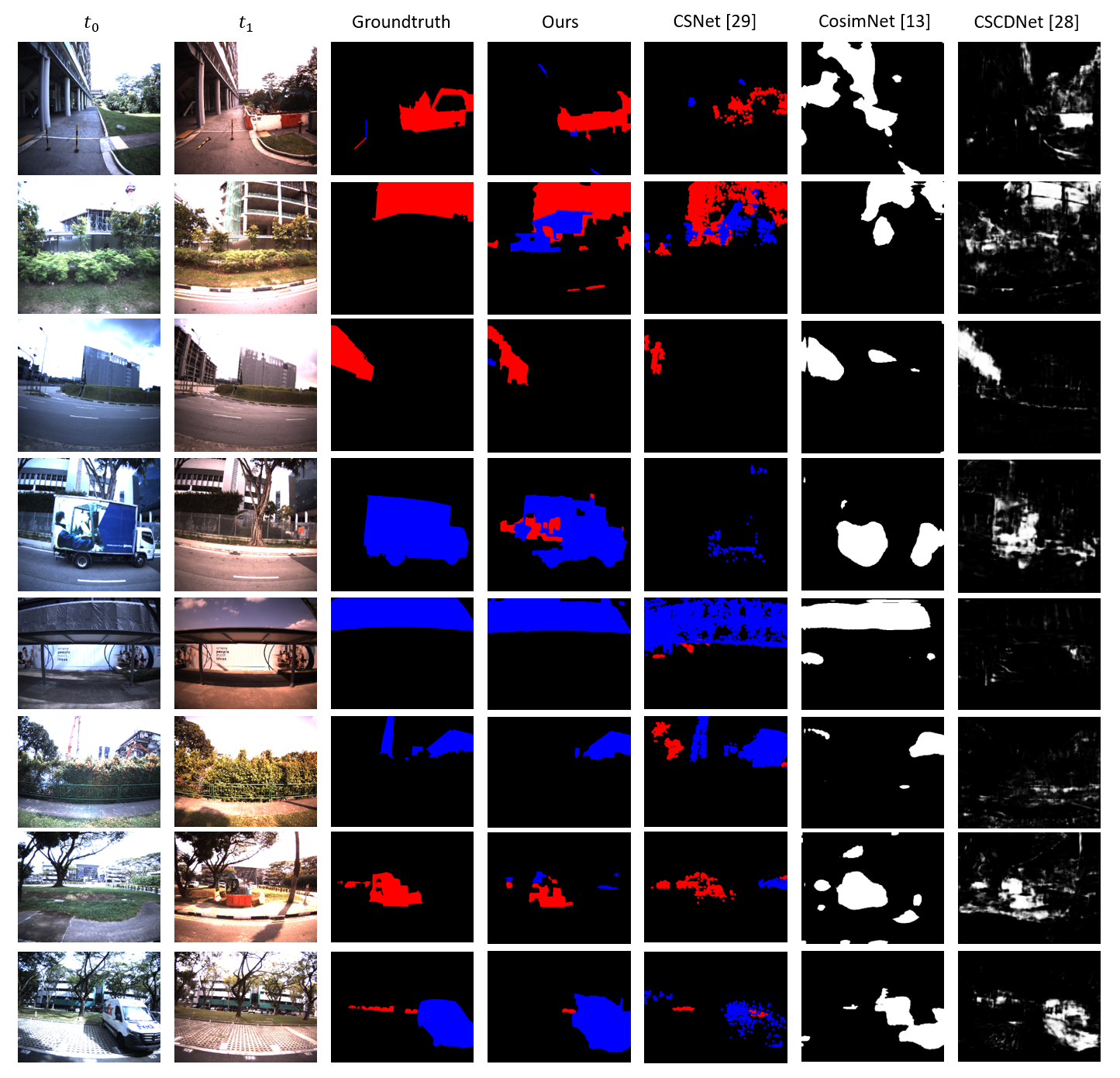}
\caption{\textbf{Change Detection Mask comparing}. Examples of change detection pipeline results, and images from the two traversals. Blue and red denote the disappearance and appearance of scene contents for pairs in times $[t_{0},t_{1}]$. While white denotes all detected changes without differentiate between appearance/disappearance contents. Rows 1-4 are from the business district, and row 5-8 are from the research town.}
\label{fig:f6}
\end{figure}

\subsection{Results} \label{sec:Results} 
\subsubsection{Map Change Detection results.}
In this section, we will examine only the QMCHANGE algorithm (See Section \ref{sec:QMCHANGE}). We compare our change detection approach to other three methods: CosimNet \cite{guo2018learning}, CSCDNet \cite{sakurada2020weakly} and CSNet \cite{yew2021city}. The evaluation datasets we used are BD and RT, same datasets used in \cite{yew2021city}. For D285 and D457 datasets QMCHANGE evaluation results, see the supplementary material, Experiments section.

We calculated three different metrics that represent the accuracy of the predicted mask - frequency weighted IOU (fwIOU),  mean IOU (mIOU), and F1-score (see the supplementary material, Semantic Change Detection metrics section).Table \ref{Tab:SCD-comparison} shows the metrics results for each dataset separately (Business District and Research Town) and for all datasets together. As shown, Our approach achieved higher performance results in all metrics. In particular, the F1 results which combined the recall and precision performance show a significant improvement of $20\%$.

The qualitative comparisons are shown in Fig. \ref{fig:f6}. In most of the examples we see our results was similar to the ground truth but there are some limitations. In some of the examples we filtered small changes (e.g the fallen cone in Fig. \ref{fig:f6} first row) or missing object that are not separated right in the visual segmentation block (e.g. the crane that blends in with the sky in Fig. \ref{fig:f6} fifth row).

\begin{table}
	\centering
	\caption{Semantic Metrics on City Scale Scene Change Detection Datasets}   
	\label{Tab:SCD-comparison}
 	\setlength{\tabcolsep}{2.8pt}
	\begin{tabular}{ |l|c|c|c|c|c|c|c|c|c| } 
		\hline
		 & \multicolumn{3}{|c|}{Business District} & \multicolumn{3}{|c|}{Research Town} & \multicolumn{3}{|c|}{(All)}\\ \hline
         Method & fwIOU & mIOU & F1 & fwIOU & mIOU & F1 & fwIOU & mIOU & F1 \\ \hline
		 CosimNet \cite{guo2018learning} & 0.5704 & 0.5781 & 0.4242 & 0.5824 & 0.5580 & 0.4319 & 0.5763 & 0.5680 & 0.4278   \\ \hline
		 CSCDNet \cite{sakurada2020weakly} & 0.5893 & 0.5800 & 0.4167 & 0.5419 & 0.5347 & 0.2889 & 0.5668 & 0.5573 & 0.3586   \\ \hline
		 CSNet \cite{yew2021city} & 0.5966 & 0.6010 & 0.4384 & 0.6328 & 0.5962 & 0.4992 & 0.6135 & 0.5986 & 0.4665   \\ \hline
		 Ours & \pmb{0.7058} & \pmb{0.6411} & \pmb{0.6567} & \pmb{0.7387} & \pmb{0.6319} & \pmb{0.6879} & \pmb{0.7205} & \pmb{0.6365} & \pmb{0.6693}   \\ \hline
	\end{tabular}
\end{table}

\subsubsection{Map Update evaluation.}
To provide a qualitative evaluation of map update based on VPS results, we have chosen $100$ map images with a large percentage of changed 3D features for both D457 and D285 datasets. Running our VPS tester and comparing the results in all modes.

We see improvement in accuracy stats when removing changed points from the map. The improvement is small because the new structure appearance is different from the map appearance, minimizing the risk of false matches.
Table~\ref{Tab:VPS-comparison} shows the value of change detection and map update process, by removing obsolete features from the map, and thus improving the VPS 3DOF accuracy.

The overall VPS test accuracy of Unchanged 3D (with removing obsolete features and their corresponding 3D points) performs better than running VPS on all 3D points (map is outdated) for both D457 and D285 datasets. For the D457 dataset, the Unchanged 3D map obtained a median location error of $0.83$ meter in the test dataset (using 1 map image), while using all old 3D points of the map, the median location error yielded accuracy of $0.9$ meter, that is a $0.07$ meter improvement. In the same manner, the Unchanged 3D map obtained the $64.8\%$ of query frames in high precision location category (0-1 meter location error, query correctly positioned), while using all old 3D points of the map, only $58.2\%$ of query frames correctly positioned, that is a $6.6\%$ improvement.

D285 is more challenging dataset than D457, beacuse it has repetitive texture, and therefore it requires more map images for PNP to overcome (using 3 map images), and in fact, it obtained more PNP-RANSAC inliers than D457 dataset. For the D285 dataset, the Unchanged 3D map obtained a median location error of $1.48$ meter in the test dataset, while using all old 3D points of the map, the median location error yielded accuracy of $1.61$ meter, that is a $0.13$ meter improvement. In the same manner, the Unchanged 3D map obtained the $37.1\%$ of query frames in high precision location category (0-1 meter location error, query correctly positioned), while using all old 3D points of the map, only $36.5\%$ of query frames correctly positioned, that is a $0.6\%$ improvement. Here, the difference between all 3D and Unchanged 3D is minor, because only 8.1\% of the 3D features are removed as changed.

\begin{table}
	\centering
	\caption{VPS Test Results on D457 and D285 Datasets}   
	\label{Tab:VPS-comparison}
 	\setlength{\tabcolsep}{0.3pt}
	\begin{tabular}{ |l|c|c|c|c|c|c| } 
		\hline
		 \textbf{Map Update Datasets} & \multicolumn{2}{|c|}{\textbf{D457}} & \multicolumn{2}{|c|}{\textbf{D285}} \\ \hline
         Stats/Data & All 3D & Unchanged 3D & All 3D & Unchanged 3D \\ \hline
         Median percentile of removed 3D points & 0\% & 26.8\% & 0\% & 8.1\% \\ \hline
		 median number of inliers & 29 & 27 & 99 & 98 \\ \hline
		 median location error [m] & 0.9 & \pmb{0.83} & 1.61 & \pmb{1.48} \\ \hline
		 \% of frames positioned in range of 0-1 [m] & 58.2\% & \pmb{64.8\%} & 36.5\% & \pmb{37.1\%}  \\ \hline
		 \% of frames positioned in range of 1-3 [m] & 27.5\% & 20.9\% & 42.9\% &  43.5\%  \\ \hline
		 \% of frames positioned in range of $>$3 [m] & 14.3\% & 14.3\% & 20.6\% & 19.4\%  \\ \hline
	\end{tabular}
\end{table}

\section{Conclusion}
We have developed an efficient and effective pipeline for scene change detection and map update. The pipeline is designed as an add-on to the regular mapping and localization pipelines and as such makes use of their existing elements: map and VPS poses, feature key-points and descriptors, 2D-3D map image coordinates, etc. The crowd-source nature of our approach – using regular VPS queries and not relying on a dedicated process, makes it very effective.\\
The main achievements of the study are:
\begin{itemize}
    \item Finding good query-map image pairs for change detection, through geometric constraints based on VPS and map poses (QMGIPS).
    \item Aligning these pairs to support local comparison of appearance, by 5DOF homography estimation, and its refined variant producing more inliers and better alignment.
    \item Robust change detection under appearance changes, using deep D2NET aligned feature comparison, combined into change masks through deep visual segmentation. This approach significantly surpassed SOTA change detection scores on publicly available datasets.
    \item Aggregating QMGIPS change masks into a small set of robust masks. Then, propagating change to all map images that exhibit these changes, using global appearance descriptors and 3D proximity.
    \item Proving the value of change detection and map update, by removing obsolete features and showing that localization is not only faster, but also more accurate.
    \item Developing own map alignment code using Median Of Scales to adjust SfM to Map scale, coarse alignment using 3D-3D RANSAC between SfM and VPS poses and finally a super-fast SOTA ICP implementation.
\end{itemize}

\section*{Acknowledgment}
The authors would like to thank our Huawei colleagues for technical support and valuable discussions. In addition, the authors declare no conflict of interest.

\bibliographystyle{unsrt}

\end{document}


\maketitle

\begin{abstract}
    This is a supplementary document to the main manuscript.
\end{abstract}


\section{Outline}
This supplementary material contains three parts. The first part, describes the low level algorithms code used in our pipeline. The second part, provides the semantic metrics used for evaluation our change detection masks. The third part, giving extra information regarding the experiments, i.e., results, figures, and configuration details.

\section{Algorithms}
\subsection{5DOF Homography Estimation}
Homography can align images under a wide range of variations. However, under QMGIPS constraints and with primarily yaw rotation (negligible tilt or pitch), perspective foreshortening is limited to the horizontal (X) image dimension, as clearly observed by buildings’ vertical parallel lines remaining vertical and parallel under perspective. 


With these assumptions, we can use a degenerate case of homography, one with 5DOF (setting $f$=1 in Equation.~\ref{uv_eq_5dof}).
\begin{align}\label{uv_eq_5dof}
  u = \frac{a\cdot x+b}{e\cdot x+f} \hspace{1cm}
  v = \frac{c\cdot y+d}{e\cdot x+f}
\end{align}
Given at least 3 matched point pairs, the least squares solution for the 5 parameters is solved by Equation.~\ref{least_sqaure_sol}.
\begin{align} \label{least_sqaure_sol}
\left[ \begin{matrix}
   {{x}_{1}} & {1} & {0} & {0} & {-{x}_{1} \cdot {u}_{1}}  \\
   {{x}_{2}} & {1} & {0} & {0} & {-{x}_{2} \cdot {u}_{2}}  \\
   {{x}_{3}} & {1} & {0} & {0} & {-{x}_{3} \cdot {u}_{3}}  \\
   \vdots  & \vdots  & \vdots & \vdots  & \vdots   \\
   {0} & {0} & {{y}_{1}} & {1} & {-{y}_{1} \cdot {v}_{1}}  \\
   {0} & {0} & {{y}_{2}} & {1} & {-{y}_{2} \cdot {v}_{2}}  \\
   {0} & {0} & {{y}_{3}} & {1} & {-{y}_{3} \cdot {v}_{3}}  \\
   \vdots  & \vdots  & \vdots & \vdots  & \vdots  \\
    \end{matrix} \right] \cdot 
    \left[\begin{matrix}
       {a}\\
       {b} \\
       {c} \\
       {d} \\
       {e} \\
    \end{matrix}    
    \right]=\left[ \begin{matrix}
       {u}_{1}\\
       {u}_{2} \\
       {u}_{3} \\
       \vdots \\
       {v}_{1} \\
       {v}_{2} \\
       {v}_{3} \\
       \vdots \\
    \end{matrix} \right]
\end{align}
This allows us to formulate the following 5DOF homography RANSAC, which uses us to 5DOF pair alignment algorithm:

\subsubsection{5DOF Pair Alignment.}
\noindent
{\it C++ Computer Program}
\begin{verbatim}
program 5DOFAlignment (3x3 perspective transform matrix)
  {Input -  image1 <vector<Point2f>>: 2D points pair 1,
            image2 <vector<Point2f>>: 2D points pair 2,
            err <double>: RMSE of 5DOF Homography Coefficients};
  {Output -  a 3x3 perspective transformation matrix,
             5DOF Homography Coefficients,
             Number of RANSAC inliers, 
             sigma x,y};
   begin 
     do KNN feature matching (k ==2) between images descriptors;
     do Bidirectional features matching;
     do lowe_ratio_test, for ignore outlier matches;
     do epipolar filtering;
     calculate 5DOF Homography RANSAC;
end.
\end{verbatim}

\begin{figure}
\centering
\includegraphics[width=\textwidth]{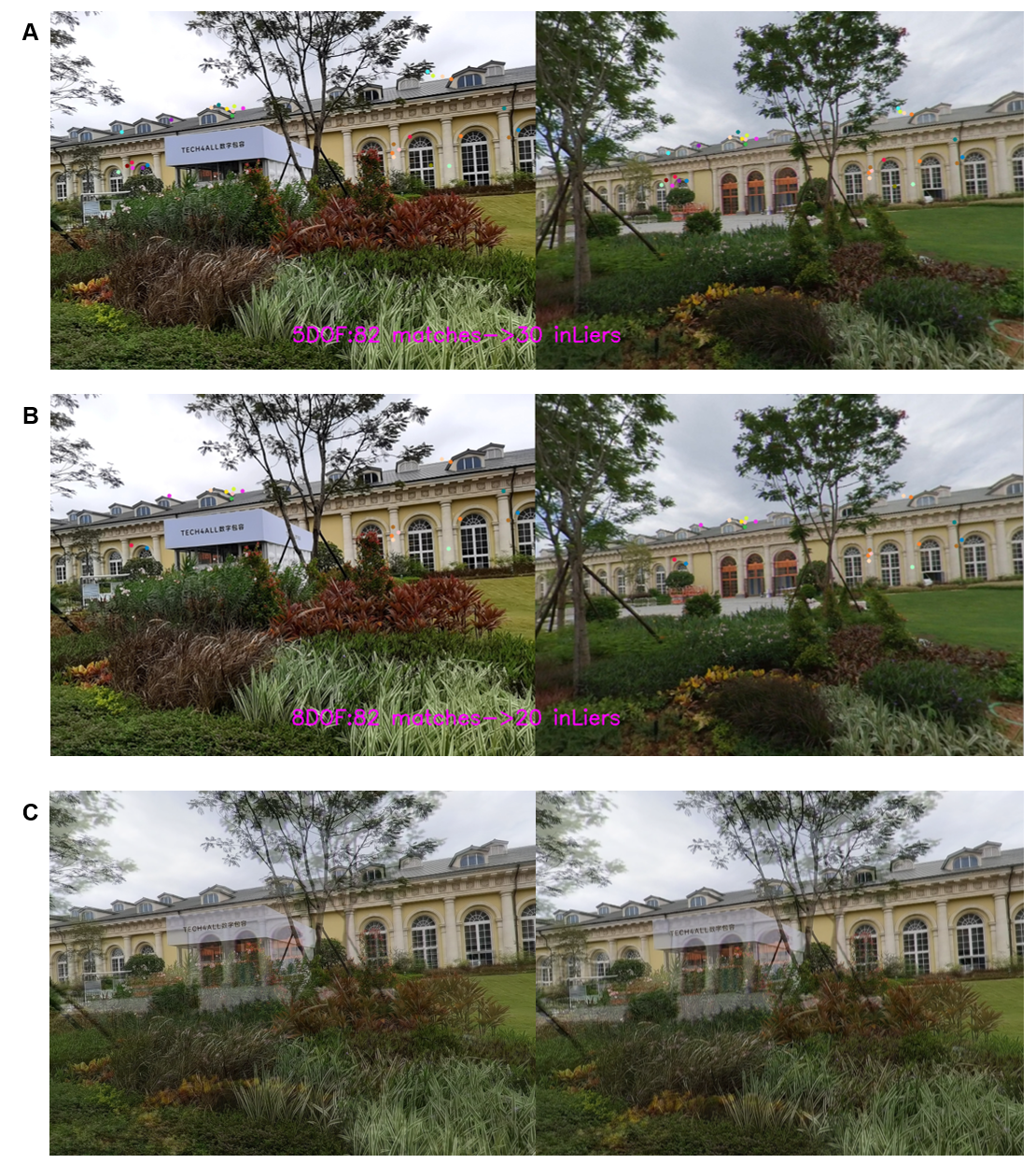}
\caption{\textbf{5DOF Homography estimation vs. 8DOF Homography estimation}. A. A query image left, map image right. 5DOF gains 30 inliers from 82 bi-directional matches. B. A query image left, map image right. 8DOF gains 20 inliers from 82 bi-directional matches. C. AddWeighted 5DOF image left, AddWeighted 8DOF image right. All relevant feature matches are along the structure (which is partly changed), with a small variance in their vertical distribution. Therefore, 5DOF which limits perspective distortion to the horizontal axes, is better.}
\label{fig:f12_sup}
\end{figure}

Yet, in several test set a significant number of map images are tilted and 8DOF performs better than 5DOF. Therefore, we run both 5DOF and 8DOF and select the one with larger number of inliers (Fig.~\ref{fig:f12_sup}).
In any case we proceed to change detection, only with QMGIPS pairs that passed alignment with a sufficient number of inliers (5DOF / 8DOF) – at least 80 inliers (a parameter). 

\subsection{SFM to Map Alignment for Map Augmentation}
Given a new map section, we align it with the old map, using 3D-3D RANSAC (below), followed Iterative Closest Point (ICP) \cite{b1}:

\subsubsection{3D-3D RANSAC Code.}
\noindent
{\it C++ Computer Program}
\begin{verbatim}
Mat ransac6DOF (Output: Tfinal)
  {Input -  corners1 <vector<Point3f>>: matched 3D points pair 1,
            corners2 <vector<Point3f>>: matched 3D points pair 2,
            nLiers <int>: number of inliers>, ...
            bLiers <boolean> point pair #i is an inlier>, ...
            nErr <float> rms error of inliers};
  {Output -  a 4x4 transformation matrix ...
             between the 3D points pairs};
   vector<Point3f>inLier1;
   vector<Point3f>inLier2;
   // RANSAC flow built around the basic algorithm 
   // for deriving 6DOF from 3 point pairs
   Mat best3Pts = ransac3DinLiers(corners1, corners2, ...
   inLier1, inLier2, bLiers, nLiers);
   if (best3Pts.empty())
        return(Mat());
   // The actual algorithm for 6DOF from 3 matched 3D point pairs
   Mat Tfinal = rigid_transform_3D(inLier2, inLier1, nErr);
end.
\end{verbatim}

\section{Semantic Change Detection metrics}
We calculated three different metrics that represent the accuracy of the predicted mask - frequency weighted IOU (fwIOU) (see Equation.~\ref{eq:fwIoU}), mean IOU (mIOU) (see Equation~\ref{eq:mIoU}), and F1-score, which is the harmonic mean of precision and recall (see Equation.~\ref{eq:f1}).

\begin{equation} \label{eq:mIoU}
mIOU = \frac{1}{n_{cl}}  \sum_{i}  \frac{p_{ii}}{s_i + \sum_{j} p_{ji}-p_{ii}}
\end{equation}

\begin{equation} \label{eq:fwIoU}
fwIOU = \frac{1}{s}  \sum_{i}  \frac{s_i n_{ii}}{s_i + \sum_{j} p_{ji}-p_{ii}}
\end{equation}

\begin{equation} \label{eq:f1}
F1 =  \frac{2TP}{2TP + FP + FN}
\end{equation}

Where $p_{ij}$ denotes the number of pixels of class $i$ classified as class j. $n_{cl}$ is the total number of classes, In this case we have two classes - mask of changes and background ($n_{cl}=2$). $s_i=\sum_{j}p_{ij}$ is the total number of pixels belonging to class i, and $s=\sum_i s_i$ denotes the number of all pixels.\\ Additionally, for F1-score definition,  $TP$, $TN$, $FP$, $FN$ are classification elements that denote true positive, true negative, false positive, and false negative of the areas of each image, respectively.

\section{Experiments}
\subsection{QMCHNAGE - additional visualization results.}
\begin{figure}
\centering
\includegraphics[scale=0.5]{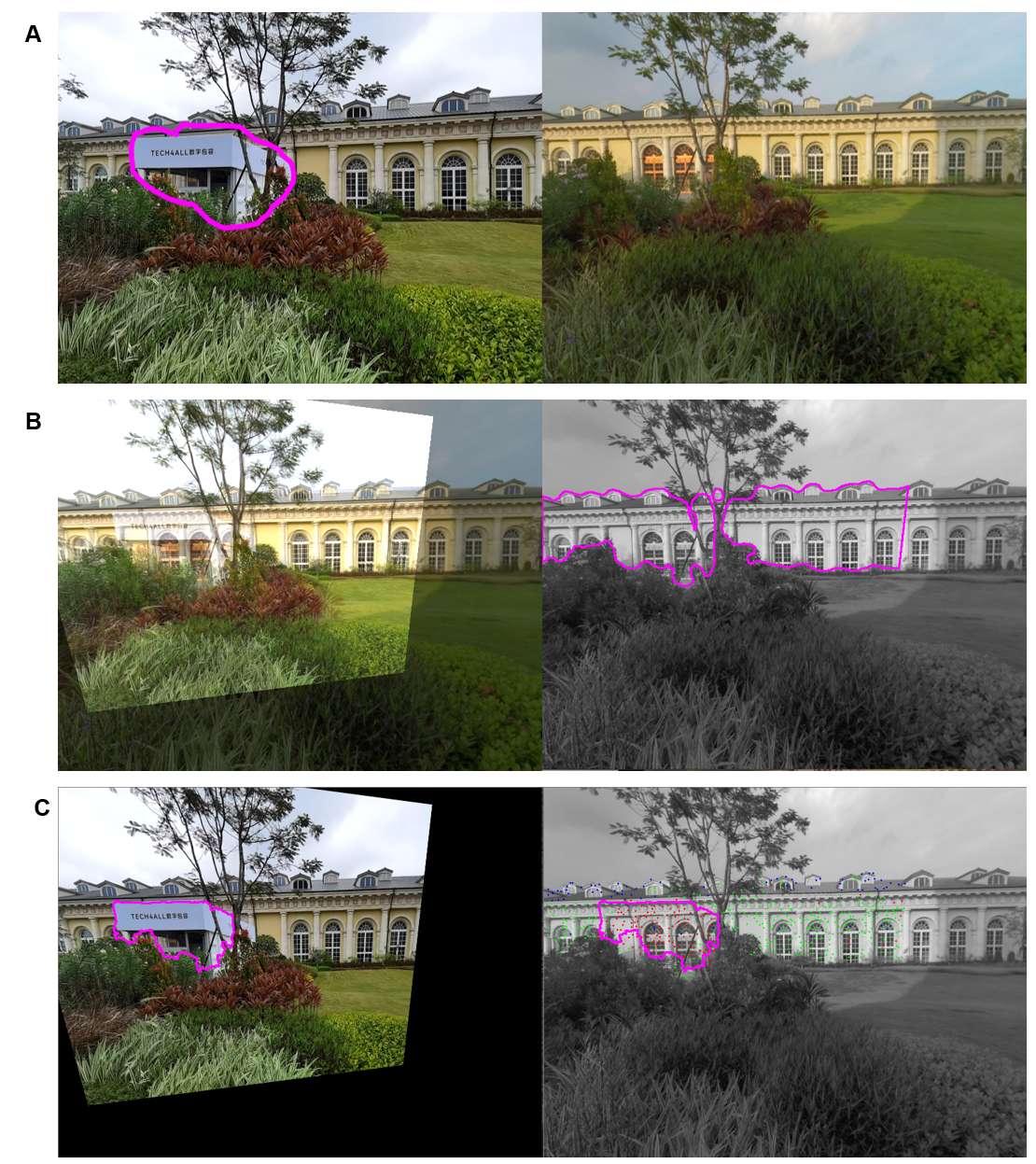}
\caption{\textbf{Change Detection algorithm results}. Examples of change detection pipeline results, and images from the two traversals. A. QMGIPS pair - query with groundtruth change (purple contour) left, map image right. B. Common area (QMALIGN) further limited by semantics (left), Map image with Query common area highlighted right (purple contour). C. Combined Changed Mask left (purple contour), map features changed (red dots) / unchanged (green dots) decision in semantically meaningful and geometrically common region right (blue dots have no 3D values).}
\label{fig:f7_sup}
\end{figure}

In this section, we show the more results of D285 dataset of our QMCHANGE algorithm. 
The qualitative result is shown in Fig.~\ref{fig:f7_sup}.

\subsection{QMPOST - detailed description.}
\begin{figure}
\centering
\includegraphics[width=\textwidth]{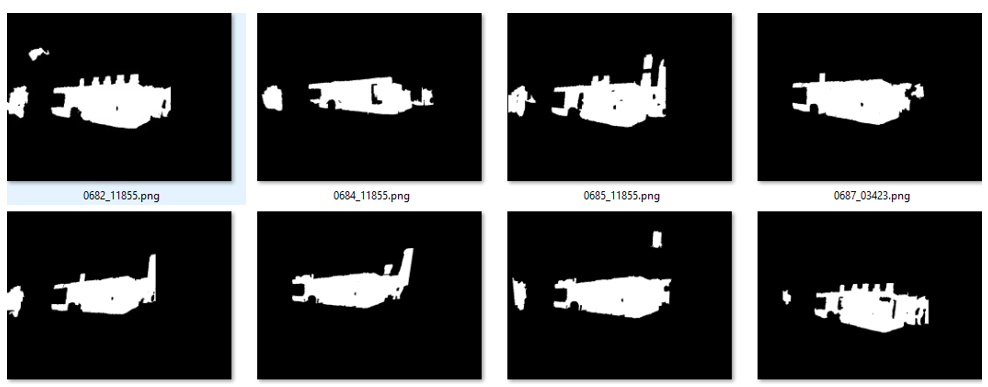}
\caption{\textbf{Noisy Individual Change Masks}. Examples of noisy change detection mask of D457 dataset.}
\label{fig:f8_sup}
\end{figure}

Query-Map image pair change masks can be quite noisy (see Fig.~\ref{fig:f8_sup}, due to a variety of reasons:
\begin{itemize}
    \item Visual segmentation \cite{b2} not necessarily aligns with change boundaries.
    \item Due to D2NET \cite{b3} feature sparseness, a matching query feature is not guaranteed to exist for each unchanged map feature. Even if exists, descriptor matching does not always produce a distance below the provided threshold.
\end{itemize}

Therefore inferring change from a single mask is not recommended. For the D457 dataset, our pairwise change detection yields $1914$ change masks. Because of the strict viewpoint proximity criteria of QMGIPS, not all changed map images have corresponding change masks.  In fact, for D457 dataset, the change masks cover $159$ map images with 1-to-85 query-map pairs each.

\begin{figure}
\centering
\includegraphics[width=\textwidth]{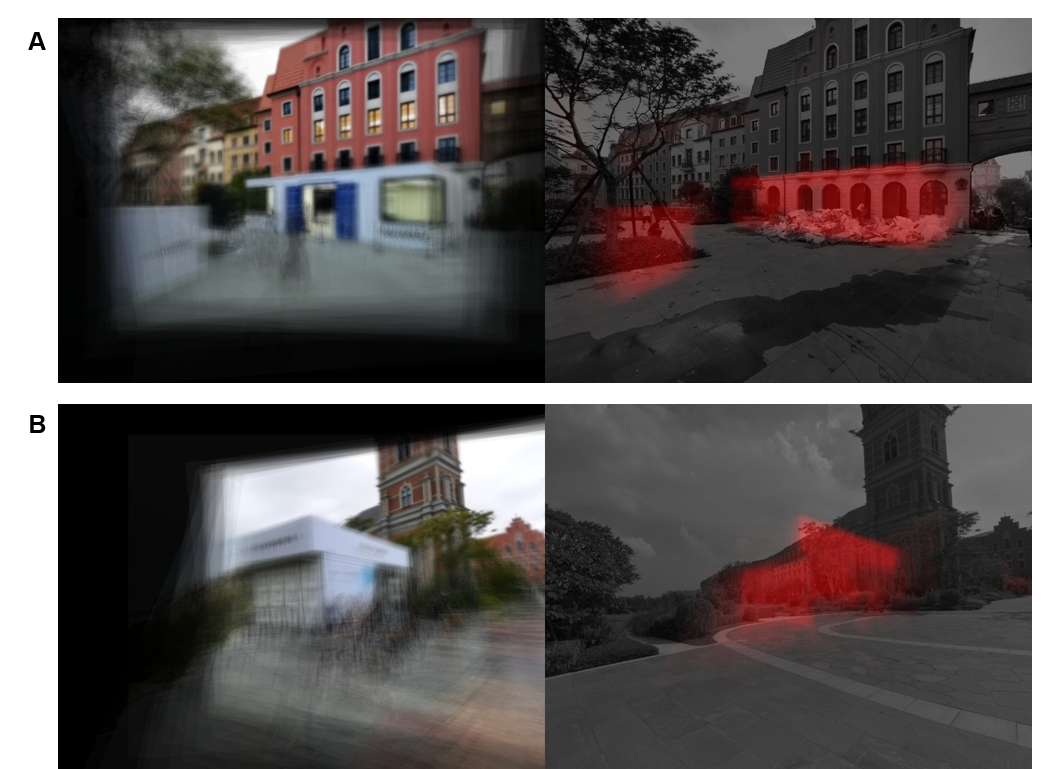}
\caption{\textbf{Change Mask Aggregation examples}. A. Master Mask (Right) and average of the aligned supporting queries (Left) – belong to D457 dataset. B. Master Mask (Right) and average of the aligned supporting queries (Left) – belong to D285 dataset.}
\label{fig:f9_sup}
\end{figure}

\begin{figure}
\centering
\includegraphics[scale=0.55]{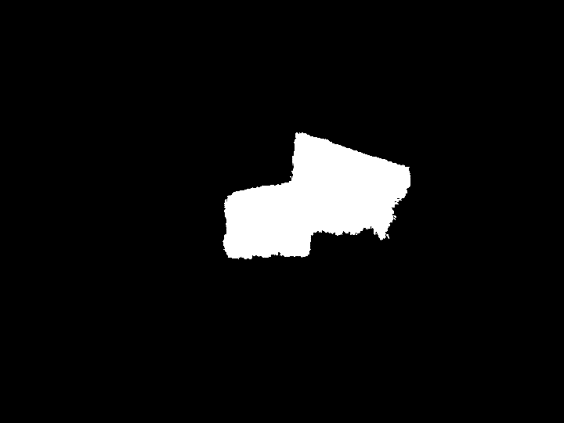}
\caption{\textbf{Final Master Change Mask}. While applying threshold to the aggregated change mask. Image binary mask is belong to D285 dataset.}
\label{fig:f10_sup}
\end{figure}

\subsubsection{Change Mask Aggregation.}	
To overcome these limitations, we aggregate multiple query-map change masks into “Master Map Change Masks”. Given a map image with at least $20$ aligned query-map pairs we can aggregate / average all change masks, aligned to that map image. Figure~\ref{fig:f9_sup}A and Figure~\ref{fig:f9_sup}B show on their right side the results of such averaging that look smooth and robust.

To visualize the actual change on the query size, we align all related query images to that same map image and average, as can be seen on the left side. To create a binary change mask that will tag changed map features, we apply a threshold to the average change mask. Therefore, if we accumulate $24$ query-map change mask, a threshold of $50\%$ means that we require a final mask pixel to appear in $12$ individual masks (see Figure~\ref{fig:f10_sup}).

\subsubsection{Indirect Change Detection.}
For the D457 dataset only $26$ map images have made is to the Master Change Mask level with $20$ or more supporting query-map features. What about the remaining map images? As mentioned above, most changed map images do not participate in QMGIPS and cannot have a change mask, let alone a master change mask.
We propose to create change lists for other map images, aiming to cover all significantly changed map image. We define a significant change when at least a certain percentage ($5\%$) of original features (with legit 3D coordinates) are changed.
We combine global similarity and common Field-of-view (FOV) to find a Master Change Mask that can be used to infer feature change indirectly. We then use 3D proximity to tag specific features as changed.
For every map image: we review all master map images, requiring:
\begin{itemize}
    \item Minimum NVLAD global descriptor \cite{b4} similarity (correlation: $\rho \geq 0.25$).
    \item Maximum FOV angular separation ($\leq 0.4$ radians).
\end{itemize}

\begin{figure}
\centering
\includegraphics[width=\textwidth]{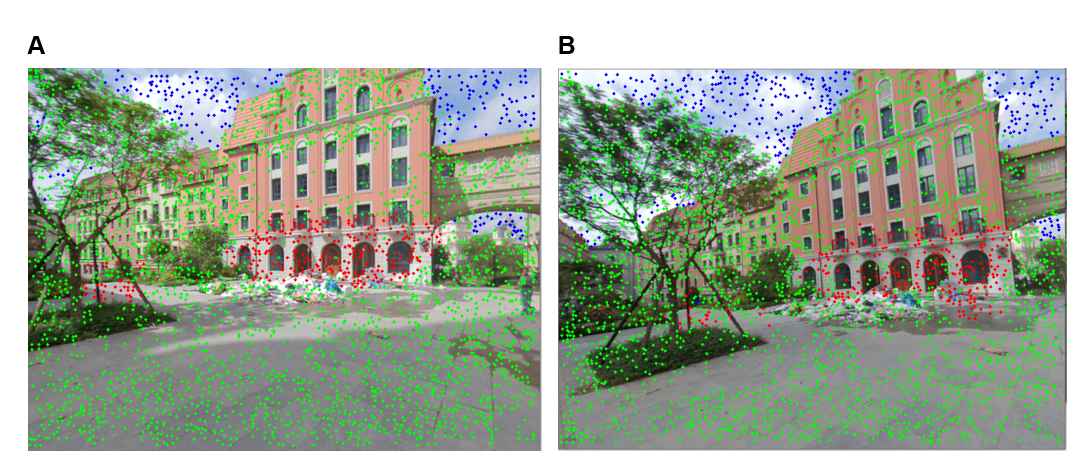}
\caption{\textbf{Indirect Change Detection} A. Master map image with changed features (in red). B. Indirectly processed map image with feature changed by proximity. Green dots indicate unchanged features, while blue dots indicate no-3D values. All Images are part of D457 dataset.}
\label{fig:f11_sup}
\end{figure}

Among these matching candidates, we take the one with highest global descriptor similarity. Figure~\ref{fig:f11_sup}A denotes the master map image, while Figure~\ref{fig:f11_sup}B shows non-master map image. For each 3D point on the image in Figure~\ref{fig:f11_sup}A, if a changed 3D point from the Figure~\ref{fig:f11_sup}B image is found within a preset search radius (1.0m) that point is tagged as changed.

\bibliographystyle{unsrt}